\pdfoutput=1

\documentclass[11pt]{article}

\usepackage[final]{ACL2023}

\usepackage{times}
\usepackage{latexsym}
\usepackage{tikz}
\usepackage{xcolor}
\usepackage{booktabs}
\usepackage{geometry}
\usepackage{tabularx}
\usepackage{booktabs}
\usepackage{xcolor}
\usepackage{listings}
\usepackage{caption}
\usepackage{tabularx}
\usepackage{etoolbox}
\usepackage{enumitem}

\usepackage{graphicx}
\usepackage{listings}
\usepackage{subcaption}
\definecolor{codebackground}{rgb}{0.95,0.95,0.95}

\definecolor{rootcolor}{RGB}{55, 55, 55}
\definecolor{reasoningcolor}{RGB}{70, 130, 180}
\definecolor{systemcolor}{RGB}{60, 179, 113}
\definecolor{planningcolor}{RGB}{244, 164, 96}
\definecolor{domaincolor}{RGB}{147, 112, 219}

\usepackage[T1]{fontenc}

\usepackage[utf8]{inputenc}

\usepackage{microtype}
\usepackage{hyperref}

\usepackage{inconsolata}

\title{TRAIL: Trace Reasoning and Agentic Issue Localization}

\author{Darshan Deshpande\quad Varun Gangal\quad
Hersh Mehta \\\textbf{Jitin Krishnan} \quad \textbf{Anand Kannappan} \quad \textbf{Rebecca Qian}\\ 
Patronus AI\\
  {\texttt{\{darshan, varun.gangal, hersh, jitin, anand, rebecca\}@patronus.ai}}
  }

\begin{document}
\maketitle
\begin{abstract}
The increasing adoption of agentic workflows across diverse domains brings a critical need to scalably and systematically evaluate the complex traces these systems generate. Current evaluation methods depend on manual, domain-specific human analysis of lengthy workflow traces—an approach that does not scale with the growing complexity and volume of agentic outputs. Error analysis in these settings is further complicated by the interplay of external tool outputs and language model reasoning, making it more challenging than traditional software debugging. In this work, we (1) articulate the need for robust and dynamic evaluation methods for agentic workflow traces, (2) introduce a formal taxonomy of error types encountered in agentic systems, and (3) present a set of 148 large human-annotated traces (\texttt{TRAIL}) constructed using this taxonomy and grounded in established agentic benchmarks. To ensure ecological validity, we curate traces from both single and multi-agent systems, focusing on real-world applications such as software engineering and open-world information retrieval. Our evaluations reveal that modern long context LLMs perform poorly at trace debugging, with the best \textsc{gemini-2.5-pro} model scoring a mere 11\% on \texttt{TRAIL}. Our dataset and code are made publicly available to support and accelerate future research in scalable evaluation for agentic workflows\footnote{\url{https://huggingface.co/datasets/PatronusAI/TRAIL}}.

\end{abstract}

\section{Introduction}

\begin{figure}[t]
    \centering
    \resizebox{0.99\columnwidth}{!}{\includegraphics{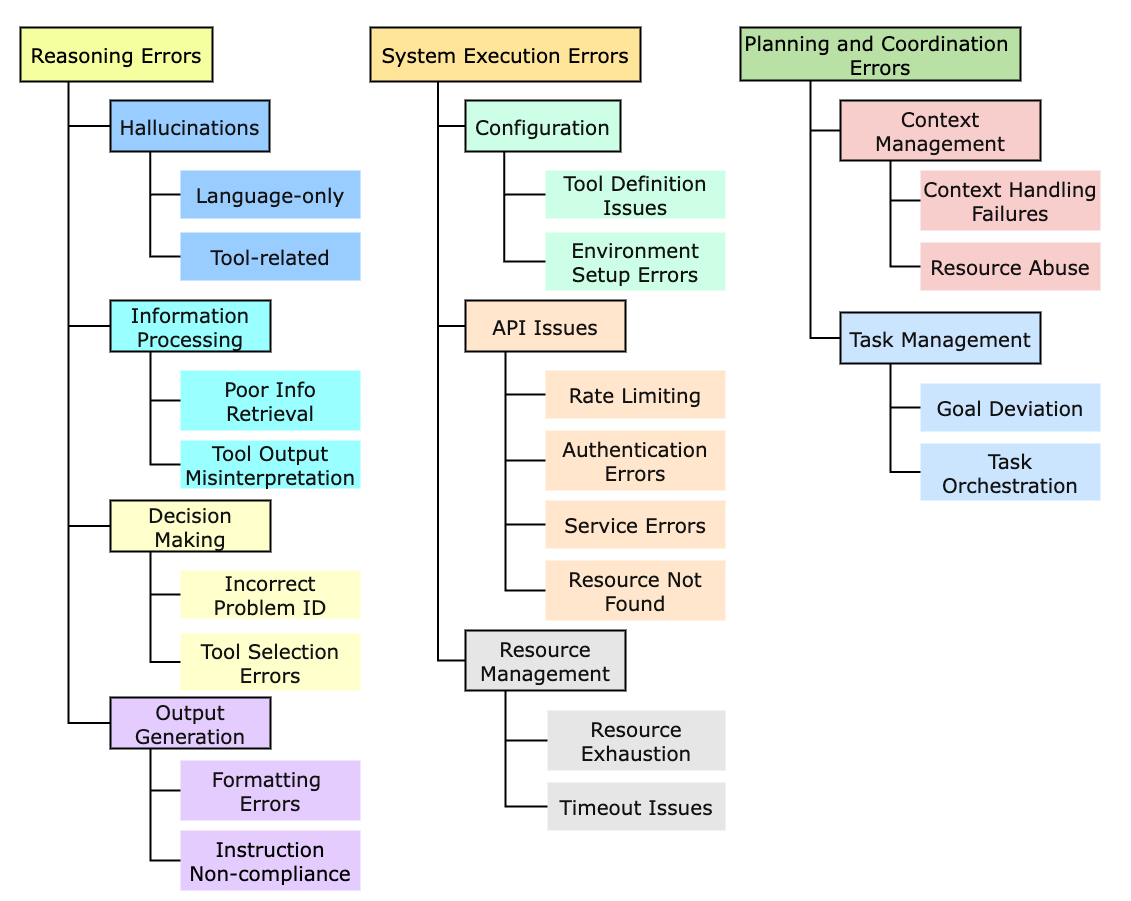}}
    \vspace{-0.4cm}
    \caption{Illustration of the TRAIL taxonomy of errors}
    \label{fig:trail_taxonomy}
    \vspace{-0.4em}
\end{figure}

The rapid advancement of large language models (LLMs) has catalyzed the development of agentic systems capable of automating difficult, multi-step tasks across various domains such as software engineering and multi-hop IR \cite{ma2023laser, OpenAI2024DeepResearch, nguyen2024dynasaur, wang2025browsing}. Unlike traditional generative models, agents can interact with diverse tools and dynamically navigate environments, often with minimal human supervision~\cite{wang2024survey}. This escalation of system complexity demands more challenging and multifaceted evaluation processes~\cite{nasim2025governance} and has led to the adoption of LLMs as evaluators for such agentic systems \cite{zheng2023judging, chen2024mllmjudge, kim-etal-2024-prometheus, zhu2025judgelm, deshpande2024glider}. 

However, as multi-agent systems scale and become integral to real-world workflows, evaluating and debugging their performance remains a significant challenge. Agentic non-determinism~\cite{laban2025llms,patronus2025statistical_risk} and multi-step task solving~\cite{mialon2023gaia,yao2024tau} demand greater observability than the simple end-to-end evaluations offered by existing benchmarks~\cite{kapoor2024ai, zhuge2024agent, moshkovich2025beyond, cemri2025multi}. Such complex environments require granular taxonomies and well-annotated traces that can serve as references for debugging and root-cause analysis of agent behaviors~\cite{cemri2025multi}.
When creating taxonomies and benchmarks to test and improve agents, we must ensure these are grounded in real-world applications and are not centered around dummy data~\cite{bowman2021will, liu2024ecbd}. Previous agent trace analysis frameworks have primarily focused on parsed traces containing unstructured text~\cite{cemri2025multi}, which do not adequately represent common agent framework outputs that generate structured traces logged in standardized formats like \texttt{opentelemetry}~\cite{opentelemetry}. Additionally, as observed by ~\citet{guo2023gpt4graph, sui2024tablellm}, handling structured data remains challenging for LLMs, an observation corroborated by previous research on automated software engineering trace analysis ~\cite{roy2024exploring, ma2024llmparser}. These limitations highlight the need for new approaches specifically designed for structured agentic traces.
To address these challenges and facilitate the analysis and evaluation of agentic executions, we propose a formal error taxonomy, shown in~\autoref{fig:annotation_categories}, that promotes granular failure diagnosis. We also present a carefully curated, turn-level annotated trace dataset called \textsc{TRAIL} (Trace Reasoning and Agentic Issue Localization), which demonstrates the validity and practical utility of our proposed taxonomy.


In our work, we utilize and build on SWE-Bench \cite{jimenez2024swebench, aleithan2024swebenchplus} and GAIA \cite{mialon2023gaia} while addressing three major shortcomings inherent to previous automatic agent evaluation paradigms. Firstly, we aim to replace end to end analysis of agents with a benchmark containing step-level analysis of traced agentic workflows. Secondly, we address the need for grounding in real scenarios by producing \texttt{opentelemetry}-based structured traces that span beyond present model context length limits. Finally, as compared to benchmarks focused only on agentic reasoning and coordination~\cite{cemri2025multi, kokel2025acp}, \textsc{TRAIL} focuses on validity through addition of finer, more aligned system execution failures and planning error categories such as \textit{API errors} and \textit{Task Orchestration Errors} to our taxonomy. Such categories are not only relevant to model developers but also to users and engineers optimizing single and multi-agent AI applications. 
The contributions of our work are as follows:
\begin{itemize}[itemsep=1mm, parsep=0.5pt]
    \item We introduce a formal taxonomy (\autoref{fig:trail_taxonomy}) that defines, fine-grained agentic error categories spanning across three key areas: reasoning, planning, and execution.
    \item Based on this taxonomy, we present \textsc{TRAIL}, an ecologically grounded execution trace benchmark comprising 148 meticulously curated traces (totaling 1987 open telemetry spans, of which 575 exhibit at least one error) drawn from the GAIA \cite{mialon2023gaia} and SWE-Bench \cite{jimenez2024swebench} datasets and covering a wide range of tasks.
    \item We show that \textsc{TRAIL} is a non-trivially difficult benchmark for LLMs on many fronts
    \begin{enumerate}
        \item Current SOTA LLM families such as \textsc{o3}, \textsc{claude-3.7-sonnet} and \textsc{gemini-2.5-pro} perform modestly at best on \textsc{TRAIL}, both in terms of predicting error categories and their location. With \textsc{gemini-2.5-pro} the best performing model, achieving only 11\% combined joint accuracy on both splits.
        \item Solving \textsc{TRAIL} requires a significant fraction of the maximum input length of LLMs (or exceeds it), as well as requires generating  significant fraction of their maximum output (See \autoref{tab:token_length_analysis}, Figure \ref{fig:length_distributions})
        \item Models benchmarked on \textsc{TRAIL} benefit from both the presence and greater extent of reasoning chains (\S \ref{subsubsec:reason_vs_nonreason}, \S \ref{subsubsec:reasoning_effort}), highlighting the need for improvement in exploration capabilities of LLMs.
    \end{enumerate} 
    \item \textsc{TRAIL} is fully open-source (MIT License), will be accompanied by a HuggingFace leaderboard, and serves as a foundation for future research on evaluating agentic workflows.
\end{itemize}


\begin{figure*}[!ht]
    \centering
    \includegraphics[width=\textwidth]{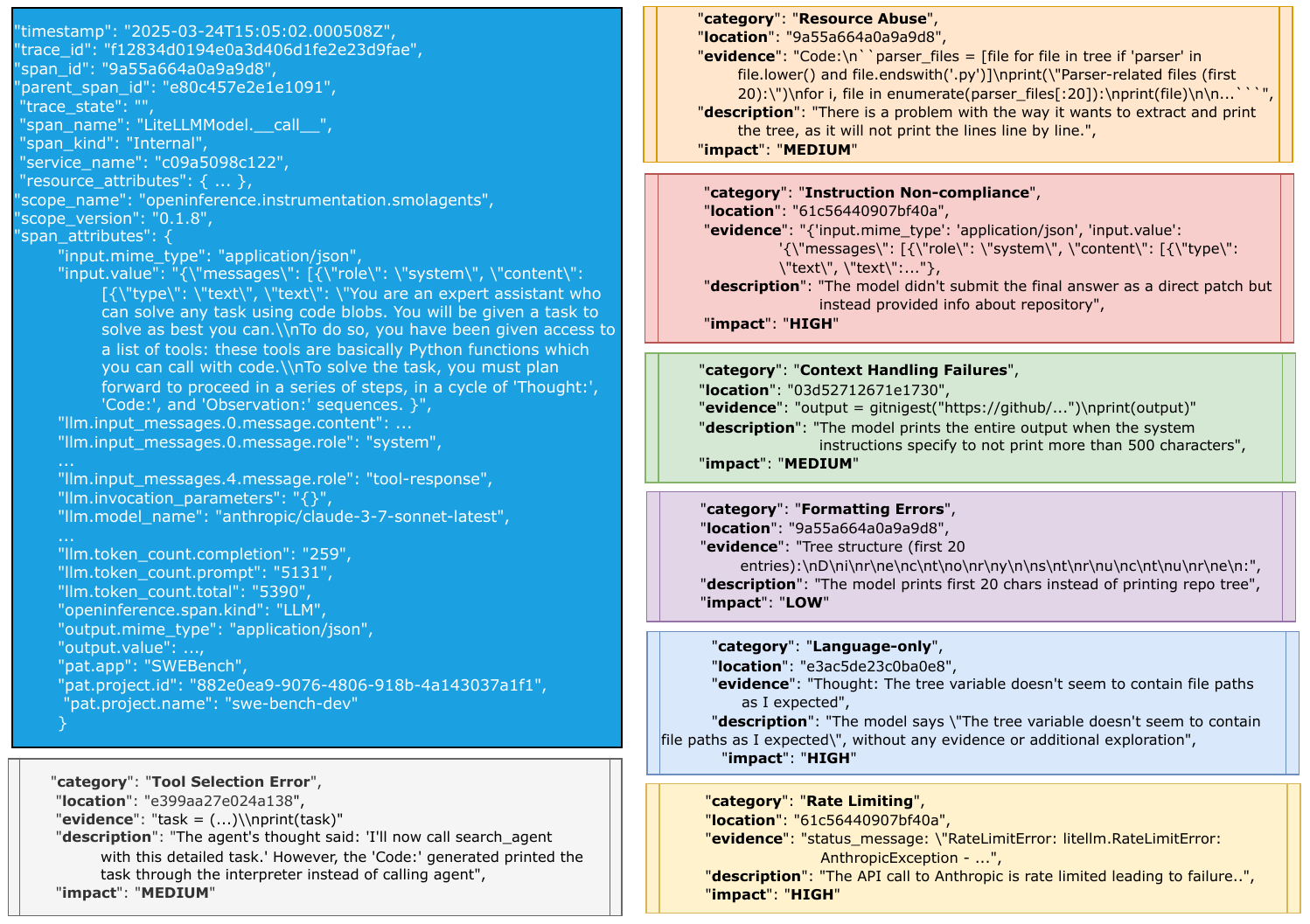}
    \vspace{-0.7em}
    \caption{TRAIL trace's span structure and error examples}
    \label{fig:trace_example_mainpaper}
    \vspace{-0.8em}
\end{figure*}

\section{Relevant Work}


\paragraph{LLM-as-a-Judge} 
Shortcomings of conventional metrics such as ROUGE, BLEU, and BERTScore~\cite{schluter-2017-limits, freitag-etal-2020-bleu, hanna-bojar-2021-fine} has led to the wide adoption of LLMs as evaluators and critics of other AI systems \cite{zheng2023judging, zhu2025judgelm, chen2025judgelrm, chen2024mllmjudge, kim-etal-2024-prometheus}. Recent approaches have enhanced LLM judges' reasoning capabilities through techniques like unconstrained evaluation plan and specialized training methods that enable more robust evaluation performance across diverse scenarios~\cite{lightman2023lets, wang2024helpsteer2, trivedi2024selfrationalization, saha2025learning}. The evaluation landscape has evolved significantly with the introduction of frameworks like FLASK~\cite{ye2023flask} which decompose coarse-level scoring into skill set-level evaluations for each instruction, demonstrating high correlation between model-based and human-based evaluations. The Prometheus models \cite{kim2023prometheus, kim-etal-2024-prometheus, kim-etal-2025-biggen} established a significant benchmark by creating judge models that surpass GPT-4 in ranking for subjective evaluation criteria. Their research also examined how performance deteriorates as subjectivity increases. More recently, several studies have enhanced judge model performance through external augmentations and checklists, highlighting the importance of incorporating high-quality reasoning chains and human guidance in model training \cite{lee2025checkeval, deshpande-etal-2024-contextualizing, deshpande2024glider, chen2025judgelrm, wang2025mcts}. Despite promising advancements, LLM judges have shown issues with propagation of biases and lack of robustness to longer inputs~\cite{ye2024justice, hu2024rethinking, wei2024systematic, zhou2025evaluating}. Since trace evaluation requires robust reasoning over large contexts~\cite{tian-etal-2024-debugbench}, LLM judges have not seen wide application in this sector yet.


\paragraph{Agentic Evaluation}

LLM-powered agents have gained significant traction for their capacity to manage intricate, sequential tasks while adaptively engaging with varied environments, rendering them particularly valuable for practical real-world applications such as software engineering and multi-hop IR~\cite{ma2023laser, OpenAI2024DeepResearch, nguyen2024dynasaur, wang2025browsing, jimenez2024swebench, qian2023chatdev, openhands, patil2024gorilla}. However, the performance gains of multi-agent frameworks remain minimal compared to their single-agent counterparts~\cite{xia2024agentless, kapoor2024agents}. As these agentic systems become more prevalent, evaluation frameworks (as compared to LLM evaluation) must offer greater customization and granularity to effectively assess the complex and sometimes unpredictable interactions between multiple agents, enabling users to precisely identify and diagnose errors at each step of the process \cite{roy2024llmrca, akhtar2025llm, jiang2025l4, zhuge2024agent, openmanusrl2024}.

\paragraph{Agent Benchmarks} 

Software engineering domain has become a fertile testbed for LLM-based collaborative problem solving for real-world use cases and to evaluate agents' ability to handle realistic coding tasks. SWE-Bench \cite{jimenez2024swebench, aleithan2024swebenchplus, pan2024swegym} was introduced as a grounded benchmark asking whether LLMs can resolve real-world GitHub issues.  
Similarly, GAIA \cite{mialon2023gaia} is a benchmark for General AI Assistants featuring real-world questions requiring reasoning, tool use, and multimodality. AssistantBench \cite{yoran-etal-2024-assistantbench} introduces a challenging benchmark of realistic, time-consuming web tasks to evaluate web agents. For agents, it is key to distinguish input sample failures from the judge model’s own internal reasoning failures. Highlighting spans can help models focus and avoid losing context while also providing additional explainability and performance improvements \cite{lv2024coarse, li-etal-2024-spotting}. Other core benchmarks include DevAI~\cite{zhuge2024agent}, MLE-bench~\cite{Chan2024MLEbenchEM}, HumanEval~\cite{du2024humaneval}, and MBPP~\cite{odena2021MBPP}. 

\paragraph{Traces and Error Taxonomies} Emerging work has emphasized the need for better observability in the agent execution traces to diagnose and manage the non-deterministic nature of agentic systems \cite{kapoor2024ai, zhuge2024agent, moshkovich2025beyond, cemri2025multi}. 
For instance, \citet{roy2024exploring} explores using LLM-based agents to dynamically collect diagnostic information from logs and metrics using retrieval tools for root cause analysis of cloud system incidents. \citet{akhtar2025llm} surveys how LLMs are being applied to automate even log analysis in security contexts. \citet{jiang2025l4} is a log analysis framework for diagnosing large-scale LLM failures based on studying real-world training failures. \citet{Ma2024LLMParserAE} explores the potential for log parsing by proposing an LLMParser delivering comprehensive evaluations in various settings.
Once the trace errors are found, to serve as references for users to debug or conduct root cause analysis of agent behaviors, these errors require a granular taxonomy~\cite{cemri2025multi,kokel2025acp, bai-etal-2024-mt}. MAST~\cite{cemri2025multi} presents an empirically grounded failure mode taxonomy but focusing only on agentic reasoning and coordination. ACPBench~\cite{kokel2025acp}, using a synthetic dataset, focuses on atomic reasoning about action and is designed to evaluate LLM's core planning skills. Other related work includes taxonomies to evaluate multi-turn conversations \cite{bai-etal-2024-mt} and designing LLM agent framework to identify and quantify complex evaluation criteria \cite{arabzadeh-etal-2024-assessing, epperson2025interactive}.

Thus, \textsc{TRAIL} distinguishes itself through its ecological validity while comprehensively addressing both single and multi-turn systems with its granular taxonomy, particularly emphasizing critical execution and planning failure patterns.

\section{Agentic Error Taxonomy}
LLM reasoning, while having advanced significantly, remains a critical source of failures in agentic workflows~\cite{costarelli2024gamebench}. These errors span several dimensions, from flawed information generation to problematic decision-making and output production~\cite{cemri2025multi}. In this section, we define a comprehensive taxonomy (as summarized in~\autoref{fig:annotation_categories}) of agentic errors spanning three key areas of failures: reasoning, planning and coordination, and system execution.

\subsection{Reasoning Errors}
\paragraph{Hallucinations}
LLMs frequently generate factually incorrect or nonsensical content, a problem that also affects agents~\cite{huang_etal_llm_hallucination_survey, ji_etal_survey_hallucination_nlg}. \textit{Text-only} hallucinations include fabricated or ungrounded statements that conflict with real-world knowledge~\cite{ji_etal_survey_hallucination_nlg}. In contrast, \textit{Tool-related} hallucinations arise when agents invent tool outputs or misunderstand tool functions, such as fabricating results or claiming nonexistent capabilities~\cite{zhang2024toolbehonestmultilevelhallucinationdiagnostic, xu2024reducingtoolhallucination}.

\paragraph{Information Processing}
Retrieval-augmented generation, which retrieves and reasons over data relevant to a query, has become increasingly popular~\cite{hu2024ragrausurveyretrievalaugmented, gao2025synergizingragreasoningsystematic}. However, recent work~\cite{xu2025doescontextmattercontextualjudgebench, su2025brightrealisticchallengingbenchmark} shows that LLMs and agents often struggle to reason effectively over retrieved information. These issues can be grouped into two main types: poor information retrieval and misinterpretation of outputs. \textit{Poor information retrieval}~\cite{wu2024stark} can introduce redundancy and content overload~\cite{stechly2024chain}, while misinterpretation of retrieved context (\textit{Tool output Misinterpretation})~\cite{karpinska2024one, wang-etal-2024-leave} may cause errors that propagate throughout an agent’s reasoning process, leading to broader incorrectness or inefficiencies.

\paragraph{Decision Making}


Task misunderstanding at the step level often arises from ambiguous prompts, unclear instructions, or an LLM’s inability to distinguish between prompt and data instructions~\cite{zverev2024can}. Detecting such misunderstandings (\textit{Incorrect Problem ID}) requires analyzing an agent's path, which is challenging in large contexts~\cite{yuan2024back} and reliable detection of these errors is crucial for agent improvement. Furthermore, effective decision making in agent workflows also depends on selecting the appropriate tool at each step~\cite{qin2023toolllm}. Because optimal planning and tool selection reduces cost and increases efficiency~\cite{yehudai2025survey}, we place \textit{Tool Selection Error} under \textit{Decision Making}.

\paragraph{Output Generation} 
LLMs often produce incorrectly formatted structured outputs~\cite{shorten2024structuredrag, liu2024we}, which is problematic for tool calls that need precise JSON or code formatting. To capture this, our taxonomy includes \textit{Formatting Errors}. Similarly, LLMs frequently struggle following complex/ambiguous instructions~\cite{white2024livebench, heo2024llms}, hence we subcategorize \textit{Instruction Non-compliance}.

\subsection{System Execution Errors}

\paragraph{Configuration Issues}
Incorrect agentic environment configuration can cause failures and limit agent capability~\cite{hu2024agentgen}. One key issue is \textit{Incorrect Tool Definition}, as shown by \citet{fu2024imprompter}, agents can be misled by inaccurate or obfuscated tool definitions in prompts, posing security and reliability risks. Additionally, poor setup of environment variables (\textit{Environment Setup Errors}), e.g., missing API keys or incorrect file permissions, can cause unexpected failures and disrupt reasoning paths.

\paragraph{API and System Issues}
As agentic systems combine LLMs with software tools, tool usage or implementation errors can disrupt workflows. With the rise of remote tool access via protocols like MCP~\cite{anthropic2025context}, capturing and categorizing API failures is increasingly important for prompt reporting to tool developers~\cite{shen2024llm}. Runtime errors involving agentic tools remain underexplored~\cite{milev2025toolfuzz}, so we specifically include the most common API tool errors in our taxonomy: \textit{Rate Limiting} (429), \textit{Authentication Errors} (401, 403), \textit{Service Errors} (500), and \textit{Resource Not Found} (404)~\cite{liu2023agentbench}.

\paragraph{Resource Management}
Resource management is crucial for agents using operating system tools like interpreters or terminals. Poor task planning can expose vulnerabilities, such as \textit{Resource Exhaustion} from overallocation~\cite{ge2023llm} or \textit{Timeout Issues} from infinite loops~\cite{zhang2024breaking}, potentially causing memory overflows or system overloads. Early detection of these errors is vital to prevent infrastructure failures.

\subsection{Planning and Coordination Errors}

\paragraph{Context Management}
As planning and reasoning become integral to agentic workflows~\cite{yao2023react, ke2025survey_frontiers_in_agentic_systems}, agents must manage long-term context, including episodic and semantic information~\cite{zhang2024surveymemorymechanismlarge}. In our taxonomy, we categorize failures in context or instruction retention as \textit{Context Handling Failures}. Additionally, repeated tool calls~\cite{kokane2024spectool} (\textit{Resource Abuse}) reflect shortcomings in planning, context management, and tool use, which our taxonomy also captures.

\paragraph{Task Management}
Environmental misconfigurations or LLM hallucinations can distract agentic systems, and poor recovery from such distractions often leads to goal deviation~\cite{ma2024caution}. These issues are amplified in multi-agent setups with sub-tasks, making effective task orchestration crucial. Therefore, we include \textit{Goal Deviation} and \textit{Task Orchestration Errors} in our taxonomy.

\begin{figure*}[ht!]
    \centering
    \begin{subfigure}[b]{0.57\textwidth}
        \centering
        \includegraphics[width=\textwidth]{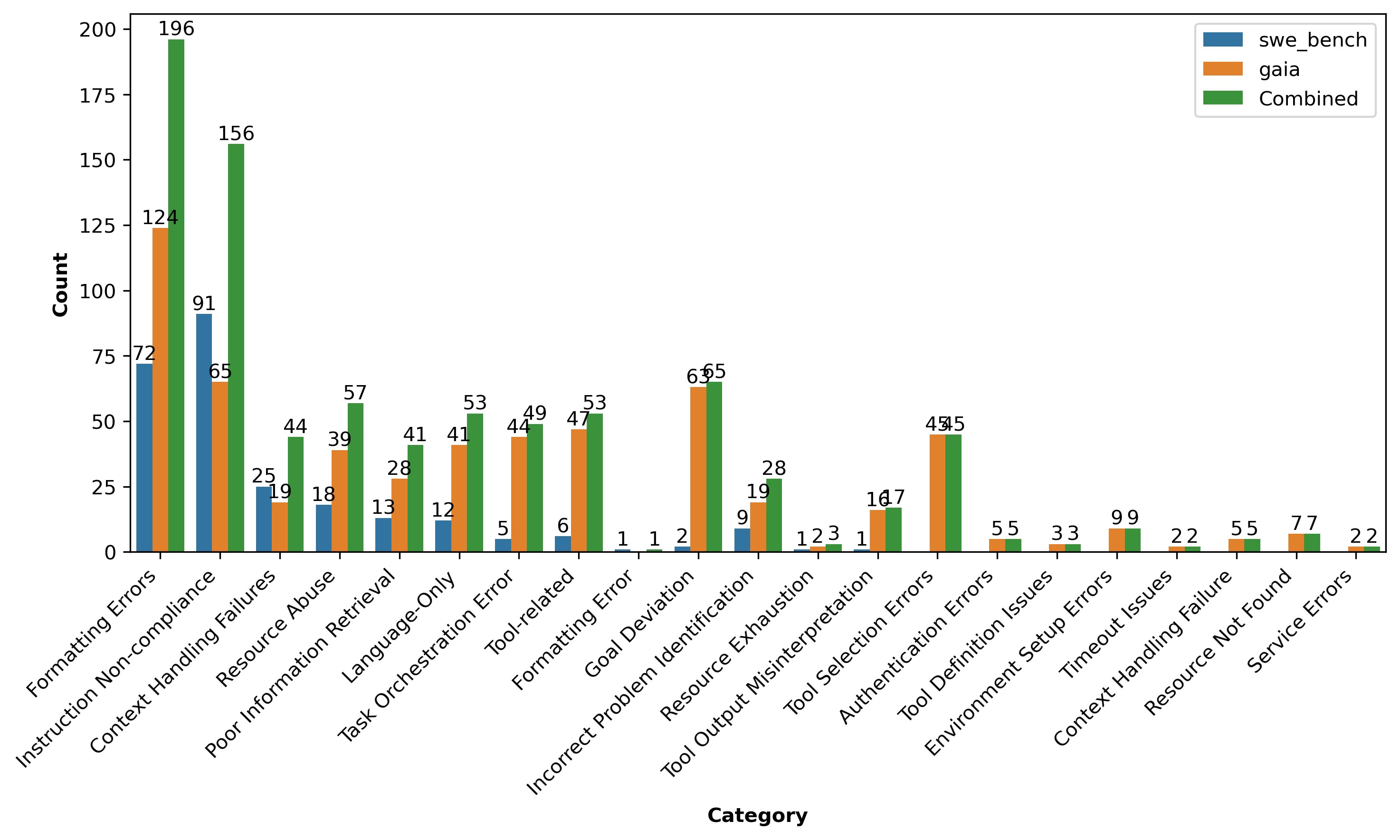}
        \vspace{-2em}
        \caption{Error Category Annotations in TRAIL}
        \label{fig:category_annotation_distribution}
    \end{subfigure}
    \hfill
    \begin{subfigure}[b]{0.42\textwidth}
        \centering
        \raisebox{4.0em}{\includegraphics[width=\textwidth]{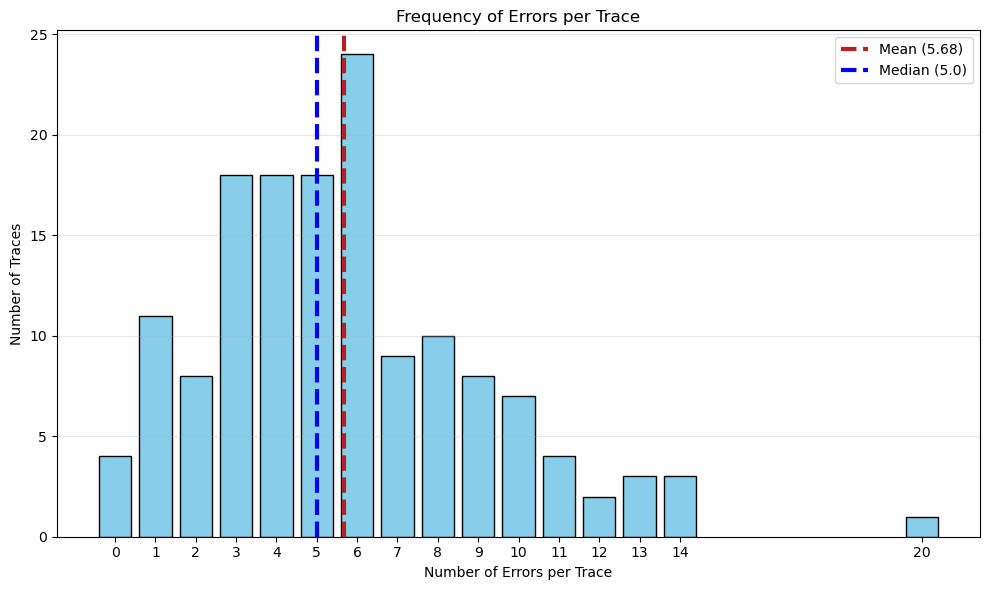}}
        \vspace{-2em}
        \caption{Distribution of Errors per Trace}
        \label{fig:trace_error_dist}
    \end{subfigure}
    \vspace{-1.2em}
    \caption{\textsc{TRAIL} Dataset Statistics}
    \label{fig:annotation_categories}
\end{figure*}

\section{TRAIL Benchmark}
\textsc{TRAIL} is a benchmark aimed to evaluate LLM capabilities to analyze and evaluate long, structured, \texttt{opentelemetry} standardized agentic executions. 
\textsc{TRAIL} follows our fine grained taxonomy and contains 148 carefully annotated agentic traces. The dataset uses text-only data instances from the GAIA~\cite{mialon2023gaia} and SWE Bench Lite~\cite{jimenez2024swebench} datasets, spanning multiple information retrieval and software bug fixing tasks. It contains a total of 841 annotated errors, averaging at 5.68 errors per trace~\autoref{fig:annotation_categories}.

\subsection{Goals and Design Choices}
\paragraph{Core Agent Task}
We aim to showcase realistic agentic workflows and so we target two widely adopted agentic datasets, the GAIA benchmark~\cite{mialon2023gaia}, an open world search task, and the SWE-Bench-Lite~\cite{jimenez2024swebench} dataset, for locating and fixing issues in Github repositories. We select these datasets due to their challenging nature and necessity for environment and search space exploration.

\paragraph{Agent Orchestration}
~\citet{liu2023bolaa} first presented a standardized hierarchical method of orchestrating agents, derivatives of which are actively adopted by several works~\cite{zhao2024epo, zhao2025cityeqa}. We closely follow this hierarchical structure and adopt the Hugging Face OpenDeepResearch agent~\cite{huggingface2024opendeepresearch} for creating traces for the GAIA benchmark. We select the state-of-the-art \verb|o3-mini-2025-01-31|~\cite{openai_o3mini} and assign it as the backbone model for the manager and search agents respectively because of its strong tool use and planning ability as showcased by~\citet{phan2025humanitysexam}. For more information, refer to \S\ref{appendix: agent_orchestration}.

Parallelly, to explore single-agent planning errors and elicit context handling errors for the SWE-Bench split, we use a CodeAct agent~\cite{wang2024codeact} and provide it access to a sandboxed environment, a python interpreter and the \verb|gitingest|\footnote{\url{https://github.com/cyclotruc/gitingest}} library. We select \verb|claude-3-7-sonnet-20250219| as the backbone model due to its strong performance on software engineering tasks~\cite{anthropic_claude_3_7_sonnet}. To further organically introduce errors into this agent system, we add instructional constraints such as output length limits and force exploration via prompts. The complete prompt is at \S\ref{appendix: swe_bench_prompts}.

\paragraph{Workflow Tracing}
To ensure compatibility of this dataset with real world tracing and observability software, all traces are collected via \verb|opentelemetry|~\cite{opentelemetry}, specifically, its most widely adopted open-source derivative compatible with agents, the \verb|openinference| standard ~\cite{openinference} as adopted by~\citet{moshkovich2025beyond}. 

\begin{table*}[!ht]
    \centering
    \small
    \setlength{\tabcolsep}{4pt}
    \begin{tabular}{l c c c c c c c c}
        \toprule
         & \multicolumn{4}{c}{\textsc{TRAIL} (GAIA)} & \multicolumn{4}{c}{\textsc{TRAIL} (SWE Bench)} \\
         \cmidrule(lr){2-5} \cmidrule(lr){6-9}
         Model & Cat. F1 & Loc. Acc. & Joint & $\rho$ & Cat. F1 & Loc. Acc. & Joint & $\rho$ \\\midrule
         \textsc{Llama-4-Scout-17B-16E-Instruct}\textsuperscript{\textdagger} & 0.041 & 0.000 & 0.000 & 0.134 & 0.050 & 0.000 & 0.000 & 0.264 \\
         \textsc{Llama-4-Maverick-17B-128E-Instruct}\textsuperscript{\textdagger} &  0.122 & 0.023 & 0.000 & 0.338 & 0.191 & 0.083 & 0.000 & -0.273  \\
         \textsc{GPT-4.1}\textsuperscript{\textdagger} & 0.218 & 0.107 & 0.028 & 0.411 & 0.166 & 0.000 & 0.000 & 0.153\\
         \textsc{Open AI o1\textsuperscript{*}} & 0.138 & 0.040 & 0.013 & 0.450 & \texttt{CLE} & \texttt{CLE} & \texttt{CLE} & \texttt{CLE} \\
         \textsc{Open AI o3\textsuperscript{*}} & 0.296 & 0.535 & 0.092 & 0.449 & \texttt{CLE} & \texttt{CLE} & \texttt{CLE} & \texttt{CLE} \\
         \textsc{Anthropic Claude-3.7-Sonnet\textsuperscript{*}} & 0.254 & 0.204 & 0.047 & \textbf{0.738} & \texttt{CLE} & \texttt{CLE} & \texttt{CLE} & \texttt{CLE} \\
         \textsc{Gemini-2.5-Pro-Preview-05-06}\textsuperscript{*\textdagger} & \textbf{0.389} & \textbf{0.546} & \textbf{0.183} & 0.462 & 0.148 & \textbf{0.238} & \textbf{0.050} & \textbf{0.817} \\
         \textsc{Gemini-2.5-Flash-Preview-04-17}\textsuperscript{*\textdagger} & 0.337 & 0.372 & 0.100 & 0.550 & \textbf{0.213} & 0.060 & 0.000 & 0.292\\
         \bottomrule
    \end{tabular}
    \vspace{-0.2em}
    \caption{Performance across LLMs for Error Categorization \& Localization on TRAIL (GAIA) and TRAIL (SWE Bench). Models marked with \textsuperscript{*} have reasoning set to "high"; \textdagger\ indicates 1M+ token context window. Insufficient context length is marked as \texttt{CLE}. Pearson correlation b/w overall human and generated scores is shown under $\rho$.\protect\footnotemark}
    \vspace{-0.2em}
    \label{tab:trail_performance}
\end{table*}

\subsection{Data Annotation and Validation}
We selected four annotators with expertise in software engineering and log debugging to label our agent traces. To assess agreement, a separate set of 63 traces was assigned. Results based on these indicate high inter-annotator agreement during curation. We defer details of our complete annotation and agreement measuring processes and actual numbers from them to \S\ref{appendix_subsec:complete_data_annot_and_valid}.

\subsection{Dataset Analysis}


Following the post-annotation review, we found errors in 114 GAIA traces and 30 from SWE Bench. As shown in~\autoref{fig:annotation_categories}, these errors cover various categories, with most falling under \textit{Output Generation}. Specifically, \textit{Formatting Errors} and \textit{Instruction Non-compliance} make up 353 of 841 total errors—nearly 42\%. In contrast, \textit{System Execution Errors} are rare. This categorical imbalance highlights two important considerations for evaluating agentic pipelines. First, the prevalence of \textit{Output Generation} errors suggests that current LLM systems struggle with high-level reasoning and understanding task parameters, even with careful prompt-engineering. Second, although infrequent, errors in categories like API failures can be catastrophic and are critical to detect, as they are often difficult to recover from, unlike errors due to goal deviation or tool misinterpretation. Most errors in our data are high or medium impact (\autoref{fig:error_impact_pie}). While model hallucinations and resource management issues greatly affect agent behavior, about 44\% of \textit{Output Generation} errors are low impact (\autoref{fig:error_impact_distribution}). This underscores need for a classification scheme that includes rare but significant error types. A key feature of our taxonomy is ability to categorize well such long-tail, high-impact errors.



\subsection{Summary of Evaluation Setup}
\label{subsec:evaluation_setup_summ}
For empirically evaluating and comparing LLM performance on \textsc{TRAIL} we choose the following LLMs --- \textsc{gpt-4.1}, \textsc{o1}, \textsc{o3}, \textsc{gemini-2.5} (both \textsc{pro}+\textsc{flash}), \textsc{claude-3.7-sonnet} and \textsc{LLama-4} (both Maverick+Scout). We defer detailed discussion of more evaluation setup specifics to \ref{appendix_subsec:evaluation_setup_full}

\section{Results}

In \S\ref{subsec:qual_quant_analysis}, we analyze the research questions below:
\begin{itemize}[itemsep=1mm, parsep=0.5pt]
\item How does long context reasoning affect TRAIL performance? How many inputs exceed the LLM's context window? How does trace length impact this? We address these in \S\ref{subsubsec:input_output_context_lengths} \S\ref{subsubsec:longcontextability_correl},  and \S\ref{subsubsec:length_correl}.
\item Does TRAIL benefit from more reasoning? We explore this in \S\ref{subsubsec:reason_vs_nonreason} and \S\ref{subsubsec:reasoning_effort}.
\item Which error categories are easier to predict? Where do non-reasoning models perform notably worse? We examine this in \S\ref{subsubsec:category_level_nuances}.
\end{itemize}

\footnotetext{All reported results are an average of three runs.}

\subsection{Qualitative and Quantitative Analysis}
\label{subsec:qual_quant_analysis}

\subsubsection{Task Difficulty - Context Length and Generation Horizon}
\label{subsubsec:input_output_context_lengths}
As seen in \autoref{tab:token_length_analysis}, the distribution of raw JSON input token lengths injested to perform our task cuts close to the input context limit of several LLMs - with the maximum input trace length always being twice longer than the input length limit, and even the mean itself sometimes going over. Furthermore, even the typical output token length horizon the LLMs need to generate for the task exceeds the 1K tokens on average, with the maximum being $\approx$3.7K at the least. Besides being a significant \% of the maximum output length, this indicates the difficultly long generation horizon TRAIL needs.

\begin{table*}[!ht]
\centering
\resizebox{\textwidth}{!}{%
\begin{tabular}{l|l||l|l||c|c|c|c||c|c|c|c}
\toprule
Task & Tokenizer & Input & Output & \multicolumn{4}{c||}{Input Context Lengths} & \multicolumn{4}{c}{Output Token Lengths} \\
& & Limit & Limit & Min & Max & Mean & StdDev & Min & Max & Mean & StdDev \\
\midrule
GAIA & gpt-4.1 (=o3) & 1M & 32.77K & 20.94K & \textbf{7.50M} & 286.85K & 768.85K & 0.11K & 4.47K & 1.11K & 0.69K \\
GAIA & gemini-2.5 & 1M & 8.19K & 23.09K & \textbf{8.25M} & 313.49K & 843.53K & 0.13K & 4.95K & 1.20K & 0.75K \\
GAIA & claude-3.7 & 200K & 128K & 23.67K & \textbf{2.66M} & \textbf{262.67K} & 456.64K & 0.12K & 5.37K & 1.23K & 0.78K \\
SWEBench & gpt-4.1 (=o3) & 1M & 32.77K & 120.40K & \textbf{2.05M} & 616.92K & 473.05K & 0.11K & 3.71K & 1.71K & 0.75K \\
SWEBench & gemini-2.5 & 1M & 8.19K & 134.88K & \textbf{2.21M} & 698.09K & 552.34K & 0.13K & 4.09K & 1.88K & 0.83K \\
SWEBench & claude-3.7 & 200K & 128K & 140.16K & \textbf{2.43M} & \textbf{727.75K} & 557.86K & 0.12K & 4.17K & 1.93K & 0.87K \\
\bottomrule
\end{tabular}%
}
\caption{Input Context Lengths and Human-Annotated Output Token Lengths Across both GAIA and SWEBench Tasks and various SOTA models and their tokenizers. Input Length aggregates that exceed the limit are \textbf{highlighted}.}
\label{tab:token_length_analysis}
\end{table*}

\begin{table}[ht]
    \centering
    \small
    \label{tab:correlations}
    \begin{tabular}{l|ccc}
    \hline
    \textbf{Corr.} & \textbf{Location Acc} & \textbf{Joint Acc} & \textbf{Categ. F1} \\
    \hline
    Pearson ($r$) & -0.379 & -0.291 & -0.296 \\
\hline
Spearman ($\rho$) & -0.508 & -0.349 & -0.225 \\
\hline
\end{tabular}
\caption{Correlations b/w Input Length \& Performance}
\label{tab:correl_perf_length_gaia_o3}
\end{table}

\subsubsection{Long Context Ability and Model Performance}
\label{subsubsec:longcontextability_correl}
We compare how the models in Table \ref{tab:trail_performance} rank based on their aggregate performance on \textsc{TRAIL} vis-a-vis the relative ranking of the subsets of these models that occur on updated long-context benchmark leaderboards Longbenchv2 and fiction.live's LongContextBench \cite{bai2024longbench,fictionlivebench2025}, and notice this differs for only one model (o3 being third best rather than best on the latter). We defer the complete detail of these rankings to \S\ref{appendix_subsec:longcontext_leaderboard_ranks}. 

\subsubsection{Performance vs Input Length}
\label{subsubsec:length_correl}
We find all performance metrics to be anti-correlated with input length, as detailed in Table \ref{tab:correl_perf_length_gaia_o3}. This supports the hypothesis that longer input raw traces increase the difficulty of \textsc{TRAIL} for models.

\subsubsection{Reasoning vs Non-Reasoning Models}
\label{subsubsec:reason_vs_nonreason}
From Table \ref{tab:trail_performance}, we see all reasoning models except \textsc{o1} outperforming non-reasoning ones on both Error category F1 and Location Accuracy. On Joint Accuracy, the gap between the two families is larger --- Reasoning models other than o1 perform at 1.5-8 times the best performing non-reasoning model.

\subsubsection{Does Reasoning Effort Matter?}
\label{subsubsec:reasoning_effort}
To systematically assess the impact of reasoning extent, we experiment with the same model (\textsc{o3}) at "high," "medium," and "low" reasoning effort levels, as set by OpenAI's \textit{reasoning.effort} parameter. We find that all three metrics, including Category F1 (0.296 → 0.277 → 0.264), decrease as reasoning effort decreases. These results empirically support that \textsc{TRAIL} performance benefits from higher reasoning effort at test time, and that the superior results for reasoning models are not solely due to improved pre- or post-training (\S \ref{subsubsec:reason_vs_nonreason}). Full ablation results are in Appendix \S \ref{appendix_subsec:reasoning_effort}.

\subsubsection{Performance Across Categories}
\label{subsubsec:category_level_nuances}

\begin{figure}[t]
    \centering
    \includegraphics[width=\columnwidth]{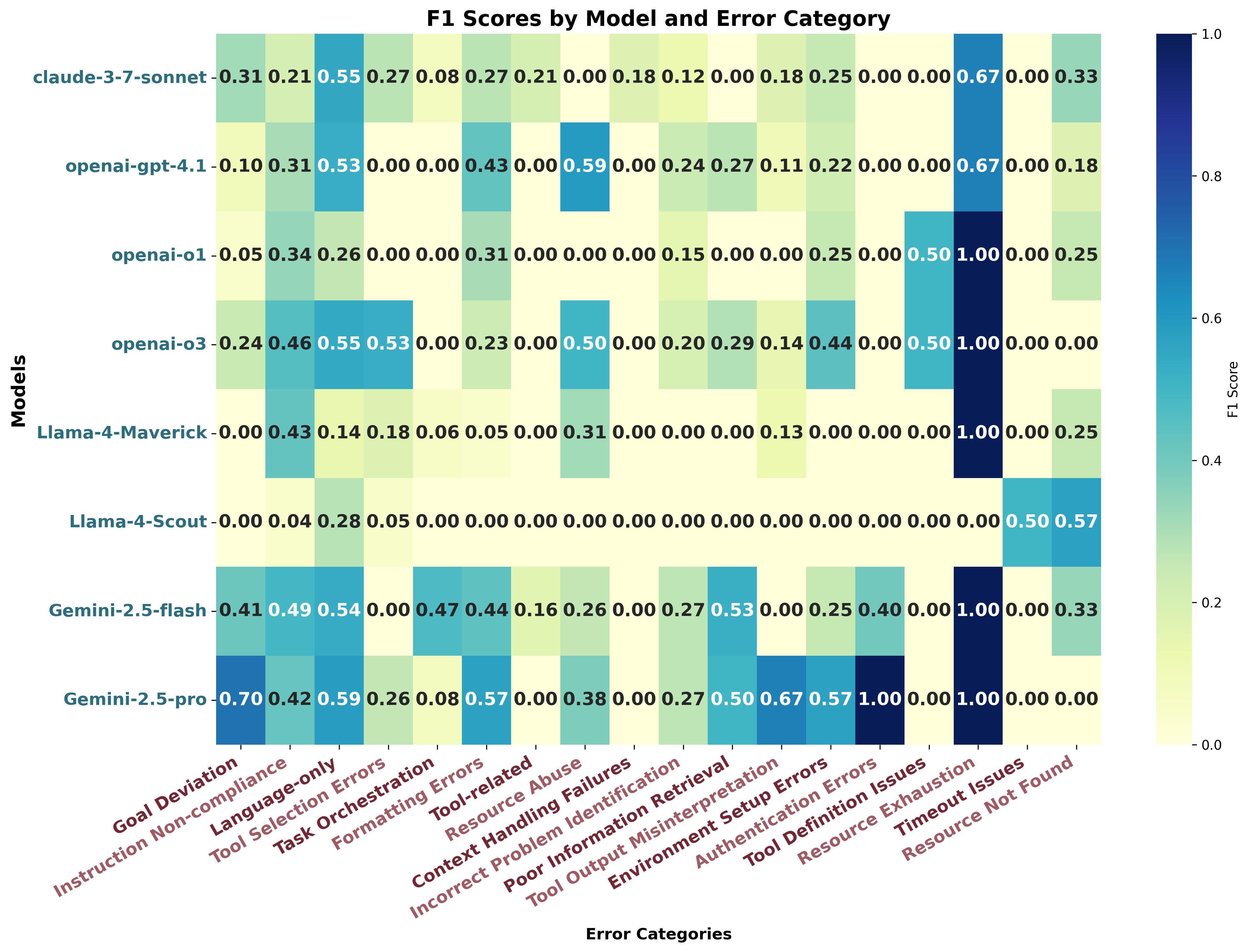}
    \vspace{-0.3cm}
    \caption{Heatmap for Error Category F1 across models; categories are ordered left to right based on their support}
    \vspace{-0.5cm}
    \label{fig:category_F1_heatmap_category_nuances}
\end{figure}

\paragraph{Hard-to-Predict Categories}
Among the most challenging categories, \textit{Context Handling Failures} stand out, as nearly all models score an F1 of 0.00, indicating these errors demand advanced reasoning. The only exception is \textsc{claude-3.7-sonnet}, which achieves a relatively better score of 0.18. \textit{Tool Selection Errors} are also difficult to predict, with most models scoring between 0.00 and 0.08, apart from \textsc{gemini-2.5-pro} (0.26), \textsc{claude-3.7-sonnet} (0.27), and especially \textsc{o3} (0.53), suggesting this is a complex error type. Similarly, \textit{Task Orchestration} shows uniformly low scores across models (0.00–0.08) except for \textsc{gemini-2.5-flash}, which stands out with a much higher F1 of 0.47.
\vspace{-0.3em}
\paragraph{Interesting Performance Divergence}
There are also categories where model performance diverges interestingly. For \textit{Goal Deviation}, \textsc{gemini-2.5-pro} and \textsc{gemini-2.5-flash} perform best (0.70 and 0.41, respectively), while \textsc{claude-3.7-sonnet} and \textsc{o3} perform moderately (0.31, 0.24); \textsc{o1} and other non-reasoning models score the lowest ($\leq$ 0.05). In the case of \textit{Poor Information Retrieval}, the two Gemini models are again notably better (0.50 and 0.53), with others at <0.30, suggesting better diagnosis of failures related to context.

\paragraph{Other Surprising Patterns}
\textit{Language-Only} errors, a subtype of hallucination, are detected relatively well by all models (0.14–0.59), implying that these are easier for models to predict even without advanced reasoning capabilities. For \textit{Formatting Errors}, performance is non-monotonic: \textsc{gpt-4.1} (0.43) and the \textsc{gemini-2.5} models (0.44–0.57) perform well, while \textsc{o1}, \textsc{o3}, and \textsc{claude-3.7-sonnet} perform worse (0.23–0.31). It is notable that \textsc{o1} and \textsc{gpt-4.1} outscore \textsc{o3} on this category, despite being older and non-reasoning respectively. We defer some model-specific observations to \S\ref{appendix_subsec:model_specific_observations}

\section{Conclusion}

In this work, \textsc{TRAIL}, a new taxonomy for classifying agentic errors, along with an expert-curated dataset of 148 agentic problem instances and 841 unique errors from GAIA and SWE Bench. Current SOTA models perform poorly as LLM Judges on this dataset, with \textsc{Gemini 2.5-pro} achieving only 18\% joint accuracy on GAIA and 5\% on SWE Bench; three out of eight models cannot even process the full context. These results highlight that existing models struggle to systematically evaluate complex agentic traces, due to the inherent complexity of agentic systems and LLM context limitations. A new framework is needed for scalable, systematic evaluation of agentic workflows.

\section*{Limitations}

The TRAIL dataset and taxonomy are primarily focused on text-only inputs and outputs but recent advancements in multimodal agentic systems require careful extension of the taxonomy to handle errors arising from new categories such as multimodal tool use. One additional limitation of TRAIL is the large number of tail categories with very few examples. It is important to ensure correctness of LLM-Judges on these categories due to the high-impact nature of the failures. Future research work can look into synthetic data generation for high-impact, low-occurrence categories by systematically modifying existing traces to induce catastrophic irrecoverable failures within the LLM context.

\section*{Ethics Statement}
While curating this dataset, we ensure that annotators are only selected based on their age (18+) and their expertise in the computer science field. Annotator selection was not based on nationality, language, gender or any other characteristic apart from these two criteria. We pay annotators a total of \$12.66 per trace where each trace takes 30-40 minutes to annotate. We ensure that the traces do not contain any PII or any explicit or biased content by manually verifying traces before forwarding these to annotators. The annotators were made aware of the open-sourcing of their work and consent was obtained beforehand.

\section*{Acknowledgments}
We would like to acknowledge industry AI practicioners: Sam Yang, Mark Klein, Pasha Rayan and Pennie Li for their feedback on our error taxonomy.  

\bibliography{custom}
\bibliographystyle{acl_natbib}

\appendix

\section{Appendix}
\label{sec:appendix}

\subsection{Prompt Structure}
\label{appendix_subsec:prompt_structure}

\subsection{Long Context Leaderboard Rankings vs TRAIL}
\label{appendix_subsec:longcontext_leaderboard_ranks}

From LongBenchv2, the rank-order \textsc{gemini-2.5-pro} > \textsc{gemini-2.5-flash} > \textsc{o1} is observed, which exactly matches the ranking we observe for these models in Table \ref{tab:trail_performance}. From fiction.live's LongContextBench, the rank order \textsc{o3} > \textsc{gemini-2.5-pro} > \textsc{gemini-2.5-flash} > \textsc{claude-3.7-sonnet} > \textsc{gpt-4.1} > \textsc{o1} > \textsc{llama4-maverick} > \textsc{llama4-scout} can be read out. Apart from the exception of \textsc{o3} being worse off than \textsc{gemini-2.5-pro} and \textsc{gemini-2.5-flash} in our case, the ranking of models for \textsc{TRAIL} matches this entirely.

\subsection{Evaluation Setup}
\label{appendix_subsec:evaluation_setup_full}
To show the effectiveness of TRAIL as a benchmark for evaluating LLM-as-judge models, we select state-of-the-art closed and open source models. For closed source models, we select OpenAI's \textsc{o1}, \textsc{o3} and \textsc{gpt-4.1} models~\cite{openai_o1, openai_o3_o4_mini, openai_gpt4_1}, Anthropic's \textsc{claude 3.7 sonnet}~\cite{anthropic_claude_3_7_sonnet} and Google's \textsc{gemini-2.5 pro} and \textsc{flash} models~\cite{google_deepmind_gemini_2025} due to their strong reasoning and agentic capabilities. For open source alternatives, we select the Llama-4 suite of models, specifically \textsc{llama-4 scout} and \textsc{maverick}~\cite{meta_llama4_multimodal} due to their long context length and good reasoning support. We use Together AI as the provider for testing Llama-4 models. We separate these open and closed models according to support for reasoning tokens and large context windows (1M+ tokens) respectively in~\autoref{tab:trail_performance}. The generation temperature and top p were set to $0$ and $1$ to maximize reproducibility for non-reasoning tests whereas we used API defaults for reasoning models.

\subsection{Reasoning Effort Ablations}
\label{appendix_subsec:reasoning_effort}
In Table \ref{tab:trail_reasoning_effort_performance_ablation} we detail the performance metrics achieved by \textsc{o3} on the GAIA split of TRAIL with different levels of reasoning effort ranging from "low" to "high", using the corresponding API parameter provided by OpenAI.
\begin{table*}
    \centering
    \small
    \setlength{\tabcolsep}{4pt}
    \begin{tabular}{l c c c}
        \toprule
         & \multicolumn{3}{c}{GAIA}  \\
         \cmidrule(lr){2-4} 
         Model & Cat. F1 & Loc. Acc. & Joint   \\\midrule
         o3 + "high" $^{*}$ & 0.296 & 0.535 & 0.092   \\
         o3 + "medium" $^{*}$ & 0.277 & 0.373 & 0.104   \\
         o3 + "low" $^{*}$ & 0.264 & 0.331 & 0.071  \\
         \bottomrule
    \end{tabular}
    \caption{Variation in performance on GAIA and SWE Bench with variation in reasoning effort}
    \label{tab:trail_reasoning_effort_performance_ablation}
\end{table*}

\subsection{Span Statistics}
\label{appendix_subsec:span_statistics}
\begin{table*}[ht]
    \centering
    \caption{Span and Error Annotation Statistics for GAIA and SWEBench Datasets}
    \begin{tabular}{l c c c c c}
    \toprule
    Dataset & Total Traces & Total Spans & Total Errors & Unique Error Spans & Error Span Total \\
    \midrule
    GAIA & 118 & 977 (mean 8.28) & 579 & 383 (3.33) & 115 \\
    SWEBench & 31 & 1,010 (32.58) & 256 & 192 (6.19) & 31 \\
    \bottomrule
    \end{tabular}
    \label{tab:dataset_span_statistics}
\end{table*}
This section details the variation in the number of input spans across \textsc{TRAIL}, both the overall spans found in the raw input trace open telemetry json files as well as the number out of these that are marked by annotators to exhibit an error.

\subsection{Model-Specific Observations}
\label{appendix_subsec:model_specific_observations}
\textsc{gemini-2.5-pro} is clearly the strongest overall, excelling particularly at \textit{Goal Deviation} (0.70), \textit{Poor Information Retrieval} (0.50), \textit{Tool Output Misinterpretation} (0.67), and \textit{Environment Setup Errors} (0.57). By contrast, \textsc{gpt-4.1} shows great variability, performing very well or moderately on some categories such as Instruction Non-compliance, \textit{Language-only}, \textit{Formatting Errors}, and \textit{Resource Abuse}, but dipping below 0.10 or even hitting zero on others, including \textit{Goal Deviation}, \textit{Tool Selection Errors}, \textit{Task Orchestration}, \textit{Tool-related} Hallucinations, and \textit{Context Handling Failures}.
\subsubsection{Visualizing Token Length Distributions}
\label{appendix_subsubsec:viztoklen}

\begin{figure*}[ht!]
    \centering
    \begin{subfigure}[b]{0.99\columnwidth}
        \centering
        \includegraphics[width=\textwidth]{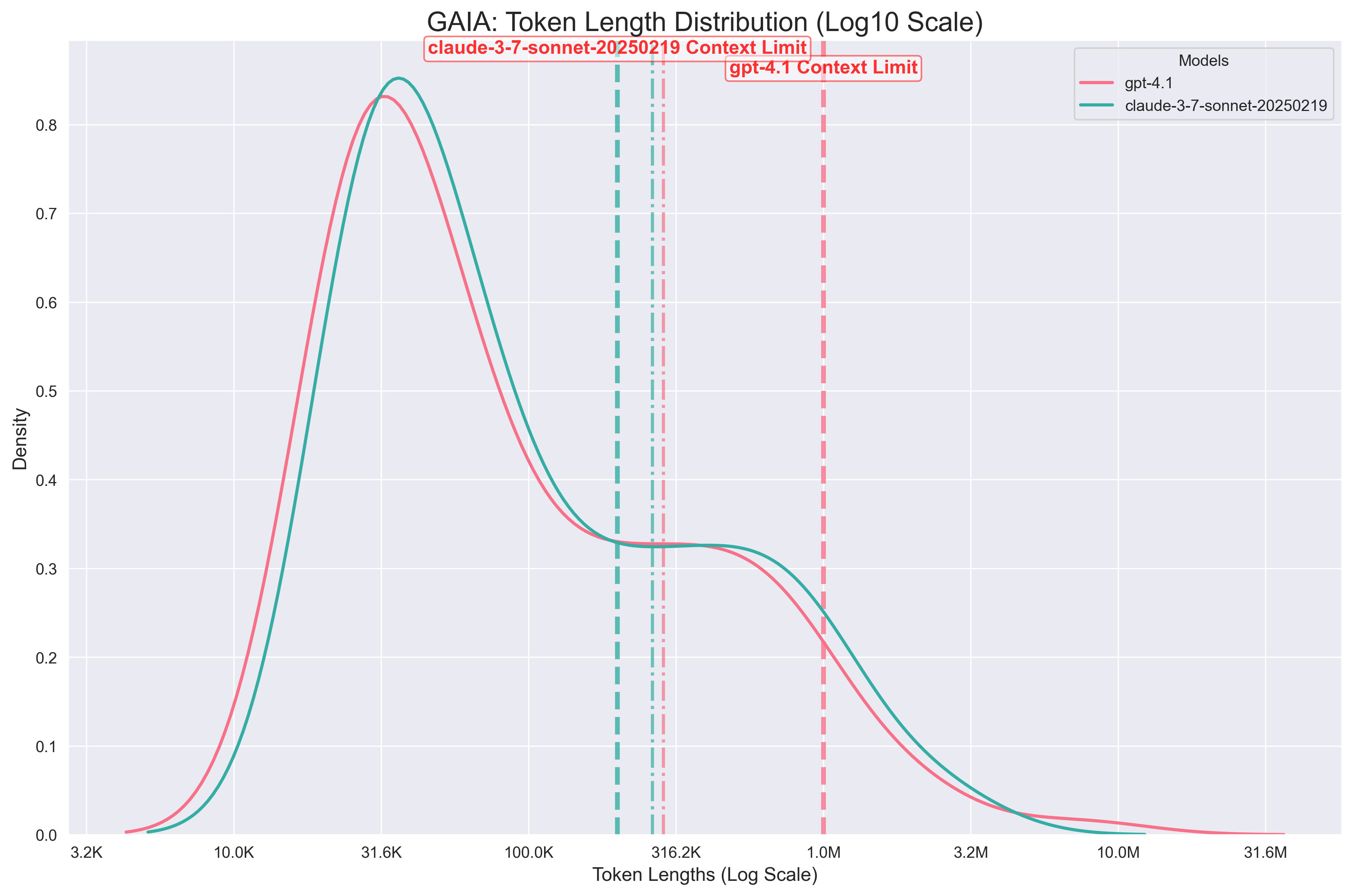}
        \caption{GAIA}
        \label{fig:input_token_lengths_gaia}
    \end{subfigure}
    \begin{subfigure}[b]{0.99\columnwidth}
        \centering
        \includegraphics[width=\textwidth]{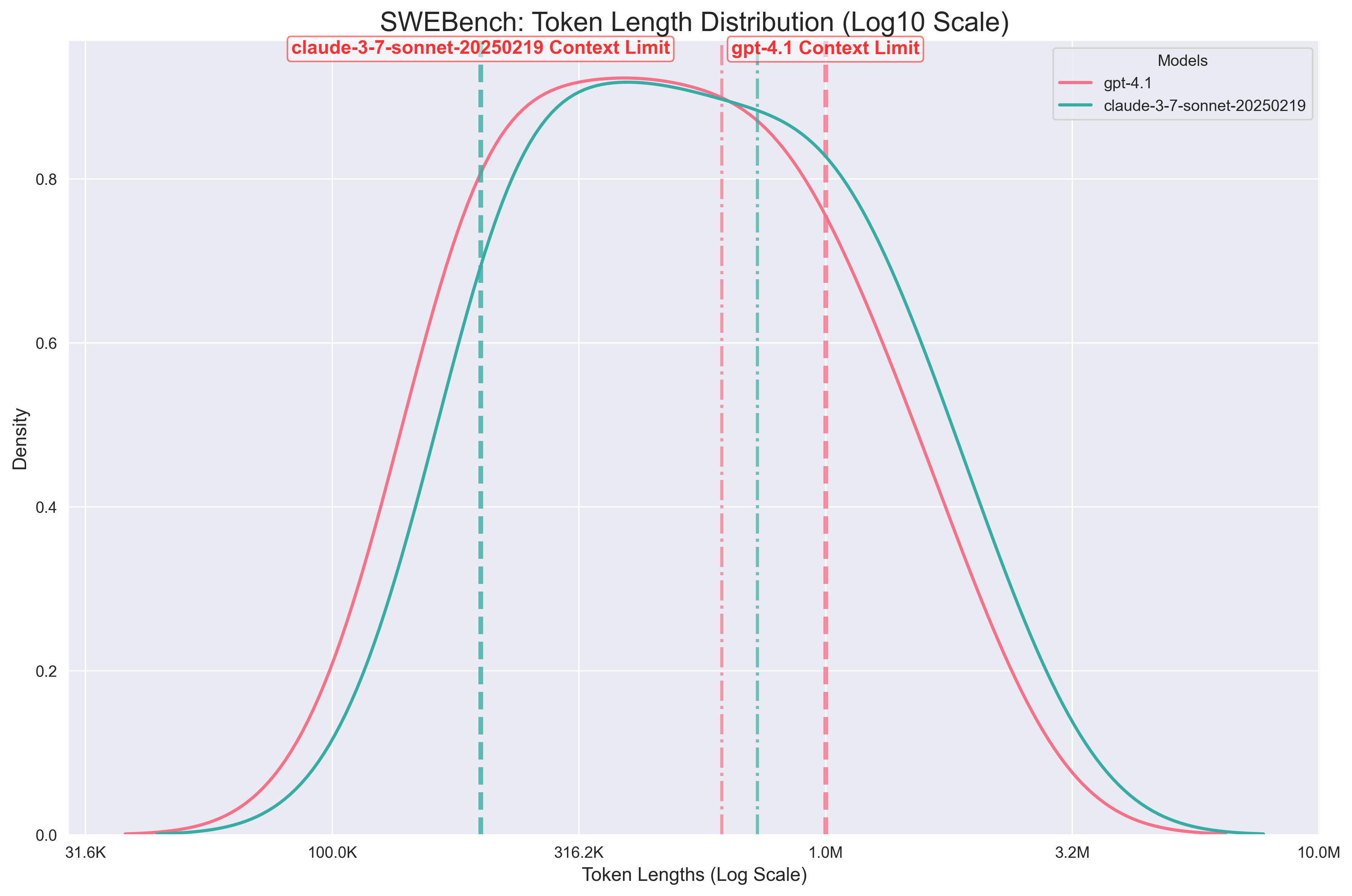}
        \caption{SWEBench}
        \label{fig:input_token_lengths_swe_bench}
    \end{subfigure}
    \caption{Input Token Length Distributions (plotted in logscale) across TRAIL tasks w.r.t  two different models for raw trace json inputs. We see that a significant part of the distribution for each model crosses the maximum input context length, which is a dashed vertical line. Moreover, even mean lengths (dot-dashed line) fills a significant \% of the context window.}
    \label{fig:length_distributions}
\end{figure*}

\subsection{Complete Data Annotation, Validation and Agreement Details}
\label{appendix_subsec:complete_data_annot_and_valid}
Due to the large trace size—often exceeding LLM context limits (\S\ref{subsubsec:input_output_context_lengths})—we conducted four independent verification rounds with ML researchers for quality assurance. Annotators evaluated each LLM and tool span in sequence, marking span ID, error category, evidence, description, and impact (Low/Medium/High) per our taxonomy, and rated overall traces for instruction adherence, plan optimality, security, and reliability (see \S~\ref{appendix:annotation_rubric}). On average, annotating a GAIA trace took 30 minutes, SWE Bench traces 40 minutes, and verification added ~20 minutes, totaling $\approx$110 and 120 minutes per trace, respectively\footnote{We did not explore and verify information (web-based or otherwise) from contents external to the trace because our baseline models are not expected to do so. Verifying such information will add more time to this estimate.}. For SWE Bench, 30 traces (444 spans) were reviewed, with 5.63\% of spans modified—mainly \textit{Resource Abuse} (33.33\%), \textit{Language-only Hallucinations} (20.83\%), and \textit{Tool-related Hallucinations} (12.5\%). For GAIA, 33 traces (697 spans) were reviewed, with 5.31\% revised, primarily \textit{Language-only Hallucinations} (23.08\%), \textit{Resource Abuse} (19.23\%), and \textit{Poor Information Retrieval} (19.23\%). These results indicate high inter-annotator agreement during curation.

\subsubsection{Annotation Rubric}
\label{appendix:annotation_rubric}
\begin{description}
    \item[\textit{Reliability score (1-5 Likert scale):}] How reliable is the agent at doing its task? Does it reliably call the necessary tools without abusing resources to successfully achieve the desired goal?
    
    \begin{itemize}[itemsep=1mm, parsep=0.5pt]
        \item[1 -] \textbf{Unreliable:} Consistently fails to utilize necessary tools, often leading to incomplete or incorrect task outcomes.
        \item[2 -] \textbf{Erratic:} Frequently misuses resources and struggles to achieve goals consistently.
        \item[3 -] \textbf{Moderate:} Sometimes meets task requirements but may inefficiently use tools.
        \item[4 -] \textbf{Reliable:} Generally effective in completing tasks with appropriate tool use.
        \item[5 -] \textbf{Highly Reliable:} Consistently achieves goals efficiently with optimal tool utilization.
    \end{itemize}
    
    \item[\textit{Security score (1-5 Likert scale):}] Did the agent make any unsafe code executions or did it try to use or access API or anything that can cause harm to the user's personal, financial or other reputation
    
    \begin{itemize}[itemsep=1mm, parsep=0.5pt]
        \item[1 -] \textbf{Very Insecure:} Regularly executes unsafe code and accesses harmful APIs, posing significant risks.
        \item[2 -] \textbf{Insecure:} Often attempts unsafe operations or API accesses that could be harmful.
        \item[3 -] \textbf{Moderately Secure:} Occasionally risky actions, but generally avoids harmful operations.
        \item[4 -] \textbf{Secure:} Rarely engages in unsafe behaviors, minimizing potential risks.
        \item[5 -] \textbf{Very Secure:} Consistently avoids unsafe code and harmful API accesses, ensuring user safety.
    \end{itemize}
    
    \item[\textit{Instruction adherence (1-5 Likert scale):}] How well was the agent able to adhere to the original task/guidelines defined by the user (first message)? Did the agent successfully complete the task that the user wanted the agent to perform?
    
    \begin{itemize}[itemsep=1mm, parsep=0.5pt]
        \item[1 -] \textbf{Poor:} Regularly deviates from instructions and fails to complete the desired task.
        \item[2 -] \textbf{Inconsistent:} Often struggles to follow guidelines and achieve the intended outcome.
        \item[3 -] \textbf{Moderate:} Sometimes adheres to instructions, but task completion is inconsistent.
        \item[4 -] \textbf{Good:} Generally follows guidelines well and completes the task successfully.
        \item[5 -] \textbf{Excellent:} Consistently adheres to instructions and successfully completes the task as intended.
    \end{itemize}
    
    \item[\textit{Plan Optimality (1-5 Likert scale):}] How well did the agent plan the task? Was it able to execute all tasks appropriately? Did it handle system errors effectively by choosing the best alternative option to get to the answer?
    
    \begin{itemize}[itemsep=1mm, parsep=0.5pt]
        \item[1 -] \textbf{Poor:} Fails to plan effectively, often executing tasks improperly and mishandling errors.
        \item[2 -] \textbf{Suboptimal:} Frequently overlooks better options, struggling with task execution and error management.
        \item[3 -] \textbf{Fair:} Adequately plans tasks with occasional missteps, sometimes handles errors.
        \item[4 -] \textbf{Good:} Plans tasks well with proper execution and effective error handling.
        \item[5 -] \textbf{Excellent:} Consistently optimal planning with efficient task execution and exemplary error management.
    \end{itemize}
\end{description}


\subsection{Correlation scores for Rubrics}
\begin{table*}[!ht]
    \centering
    \small
    \begin{tabular}{l||c c c c}
    \toprule
         Model & Reliability & Security & Instruction Adherence & Plan Optimality  \\\midrule
         \textsc{Llama-4-Scout-17B-16E-Instruct}\textsuperscript{\textdagger} & 0.09/0.25 & 1.00/1.00 & 0.075/0.08 & 0.19/0.20 \\
         \textsc{Llama-4-Maverick-17B-128E-Instruct}\textsuperscript{\textdagger} & 0.37/0.20 & 1.00/1.00 & 0.14/-0.22 & 0.33/ -0.39 \\
         \textsc{GPT-4.1}\textsuperscript{\textdagger} & 0.41/0.03 & 1.00/1.00 & 0.21/0.09 & 0.43/0.22 \\
         \textsc{Open AI o1\textsuperscript{*}} & 0.50/\texttt{CLE} & 1.00/\texttt{CLE} & 0.24/\texttt{CLE} & 0.40/\texttt{CLE} \\
         \textsc{Open AI o3\textsuperscript{*}} & 0.52/\texttt{CLE} & 1.00/\texttt{CLE} & 0.26/\texttt{CLE} & 0.44/\texttt{CLE} \\
         \textsc{Anthropic Claude-3.7-Sonnet\textsuperscript{*}} & \textbf{0.79}/\texttt{CLE} & 1.00/\texttt{CLE} & \textbf{0.53}/\texttt{CLE} & 0.59/\texttt{CLE} \\
         \textsc{Gemini-2.5-Pro-Preview-05-06}\textsuperscript{*\textdagger} & 0.59/\textbf{1.00} & 1.00/1.00 & 0.41/\textbf{1.00} & 0.15/\textbf{1.00}\\
         \textsc{Gemini-2.5-Flash-Preview-04-17}\textsuperscript{*\textdagger} & 0.58/0.61 & 1.00/1.00 & 0.39/0.12 & 0.29/0.00\\\bottomrule
    \end{tabular}
    \caption{Pearson correlation scores (GAIA/SWE Bench) between human annotators and model scores. Insufficient model context length is represented by \texttt{CLE}}.
    \label{tab:pearson_scores}
\end{table*}

As observed in~\autoref{tab:pearson_scores}, \textsc{Claude-3.7-Sonnet} receives the best scores (average of 0.738) for the GAIA subset whereas \textsc{Gemini-2.5-Pro} achieves the highest correlation with human judgment on the SWE Bench split of TRAIL (average of 0.817). 

\subsection{Distribution of Impact Levels in TRAIL instances}
\begin{figure*}
     \centering
     \begin{subfigure}[b]{0.65\columnwidth}
         \includegraphics[width=\textwidth]{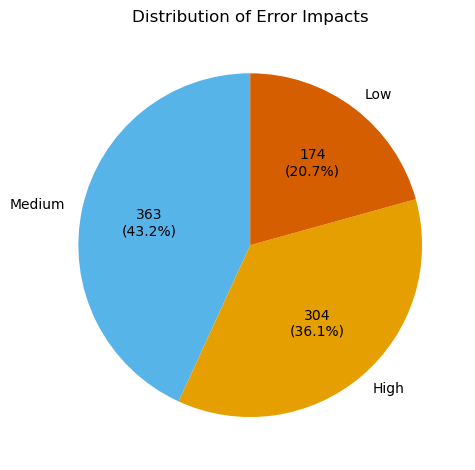}
         \caption{Error Impact Levels}
         \label{fig:error_impact_pie}
     \end{subfigure}
     \begin{subfigure}[b]{0.95\columnwidth}
         \includegraphics[width=\textwidth]{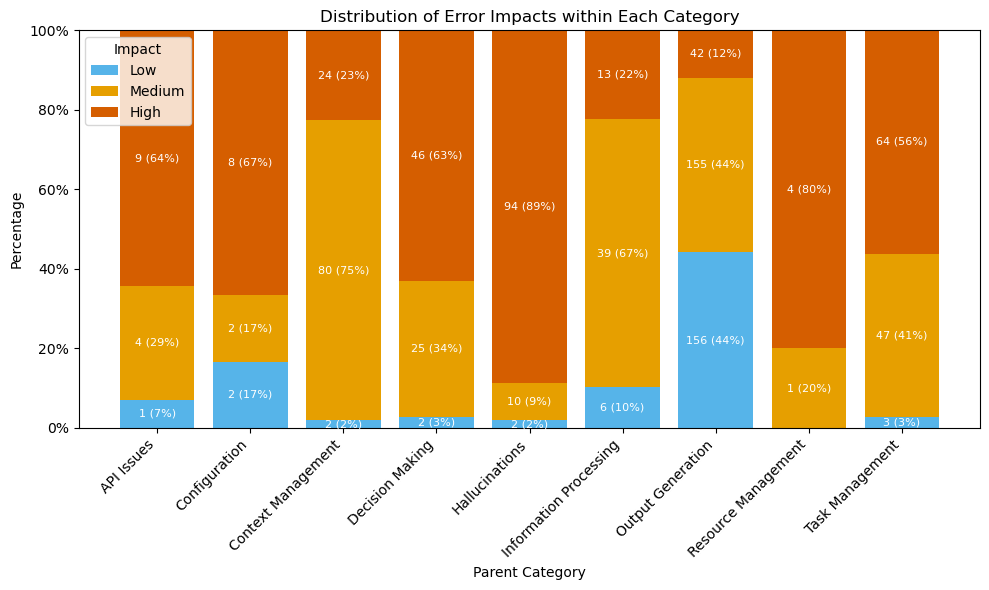}
         \caption{Impact Level of Errors for each Category}
         \label{fig:error_impact_distribution}
     \end{subfigure}
 \end{figure*}

The distribution of impact levels can be found in~\autoref{fig:error_impact_distribution}

\subsection{Agent Orchestrations for TRAIL}
\label{appendix: agent_orchestration}
\begin{figure*}
    \centering
    \includegraphics[width=0.95\linewidth]{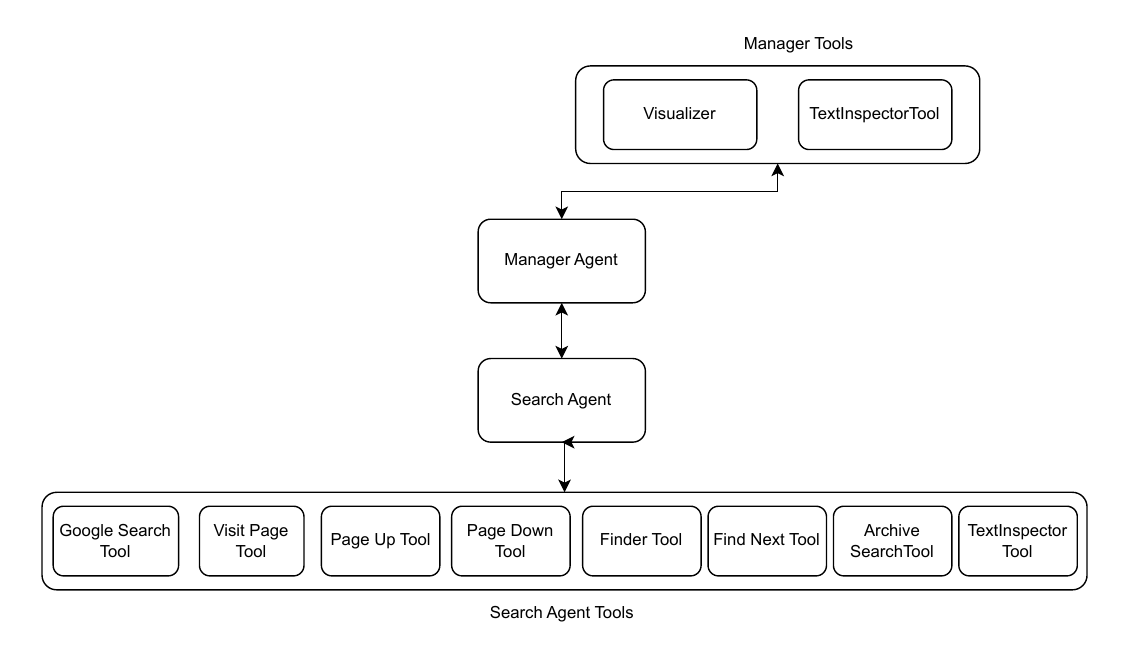}
    \caption{Search agent orchestration for GAIA dataset}
    \label{fig:search-agent-orchestration}
\end{figure*}

\autoref{fig:search-agent-orchestration} shows the agent orchestration that produces the GAIA traces. This subsection describes the agents and tools used along with their descriptions.

\paragraph{Search Agent Description}

The manager agent receives the following description for the search agent:

\texttt{
A team member that will search the internet to answer your question. Ask him for all your questions that require browsing the web. Provide him as much context as possible, in particular if you need to search on a specific timeframe! And don't hesitate to provide him with a complex search task, like finding a difference between two webpages.Your request must be a real sentence, not a google search! Like "Find me this information (...)" rather than a few keywords. }

Additional information that is provided to the search agent:

\texttt{You can navigate to .txt online files. If a non-html page is in another format, especially .pdf or a Youtube video, use tool 'inspect\_file\_as\_text' to inspect it. Additionally, if after some searching you find out that you need more information to answer the question, you can use `final\_answer` with your request for clarification as argument to request for more information.}

\paragraph{Google Search Tool}
\texttt{\
name = "web\_search"\\
description = """Performs a google web search for your query then returns a string of the top search results."""\\
inputs = {"query": {"type": "string", "description": "The search query to perform."},
"filter\_year": {"type": "integer","description": "Optionally restrict results to a certain year"}}\\
output\_type = "string"}

\paragraph{Visit Page Tool}

\texttt{\
name = "visit\_page"\\
description = "Visit a webpage at a given URL and return its text. Given a url to a YouTube video, this returns the transcript."\\
inputs = {"url": {"type": "string", "description": "The relative or absolute url of the webpage to visit."}}\\
output\_type = "string" }

\paragraph{Page Up Tool}
\texttt{\
name = "page\_up"\\
description = "Scroll the viewport UP one page-length in the current webpage and return the new viewport content."\\
inputs = {} \# This means it takes no inputs - programatically this means you call this tool as page\_up() - this is not an empty dictionary\\
output\_type = "string"}

\paragraph{Page Down Tool}
\texttt{\
\noindent name = "page\_down"\\
description = ("Scroll the viewport DOWN one page-length in the current webpage and return the new viewport content.")\\
inputs = {} \# This means it takes no inputs - programatically this means you call this tool as page\_down() - this is not an empty dictionary\\
output\_type = "string" }

\paragraph{Finder Tool}

\texttt{\
name = "find\_on\_page\_ctrl\_f"\\
description = "Scroll the viewport to the first occurrence of the search string. This is equivalent to Ctrl+F."\\
inputs = {"search\_string": {"type": "string", "description": "The string to search for on the page. This search string supports wildcards like '*'",}}\\
output\_type = "string"}

\paragraph{Find Next Tool}
\texttt{\
name = "find\_next"\\
description = "Scroll the viewport to next occurrence of the search string. This is equivalent to finding the next match in a Ctrl+F search."\\
inputs = {} \# The tool takes no inputs\\
output\_type = "string"}

\paragraph{Archive Search Tool}
\texttt{\
name = "find\_archived\_url"\\
description = "Given a url, searches the Wayback Machine and returns the archived version of the url that's closest in time to the desired date."\\
inputs = {"url": {"type": "string", "description": "The url you need the archive for."},
"date": {"type": "string","description": "The date that you want to find the archive for. Give this date in the format 'YYYYMMDD', for instance '27 June 2008' is written as '20080627'."}}\\
output\_type = "string"}

\paragraph{Text Inspector Tool}
\texttt{\
name = "inspect\_file\_as\_text"\\
description = """You cannot load files yourself: instead call this tool to read a file as markdown text and ask questions about it.
This tool handles the following file extensions: [".html", ".htm", ".xlsx", ".pptx", ".wav", ".mp3", ".m4a", ".flac", ".pdf", ".docx"], and all other types of text files. IT DOES NOT HANDLE IMAGES."""\\
inputs = {"file\_path": {"description": "The path to the file you want to read as text. Must be a '.something' file, like '.pdf'. If it is an image, use the visualizer tool instead! DO NOT use this tool for an HTML webpage: use the web\_search tool instead!", "type": "string",},
"question": {"description": "[Optional]: Your question, as a natural language sentence. Provide as much context as possible. Do not pass this parameter if you just want to directly return the content of the file.", "type": "string", "nullable": True,}}\\
output\_type = "string"}

\paragraph{Visualizer Tool}
\texttt{\
name = "visualizer"\\
description = "A tool that can answer questions about attached images."\\
inputs = {"image\_path": {"type": "string", "description": "The path to the image on which to answer the question. This should be a local path to downloaded image."},
"question": {"type": "string", "description": "The question to answer."}}\\
output\_type = "string"}

\subsection{Prompts Given to Models For Solving TRAIL}
\label{appendix_subsec:model_prompt_trail}

We give the following prompt to LLMs to generate a json with annotated error spans elements bearing location, evidence and other fields; akin to those generated in our gold annotated output jsons.
\begin{lstlisting}[
  breaklines=true,
  frame=single,
  basicstyle=\small\ttfamily,
  columns=flexible
]
Follow the taxonomy below carefully follow the instructions and provide the output in the same format as the example.

# Taxonomy
|-- Reasoning Errors
|   |-- Hallucinations
|   |   |-- Language-only
|   |   |-- Tool-related (fabricating tool outputs/capabilities)
|   |-- Information Processing
|   |   |-- Poor Information Retrieval (Tried to find information that was not relevant to the task)
|   |   |-- Tool Output Misinterpretation (Made assumptions about the tool output or used the tool output in an incorrect context)
|   |-- Decision Making
|   |   |-- Incorrect Problem Identification (Misunderstood the overall task or the local task)
|   |   |-- Tool Selection Errors (Used the wrong tool for the task)
|   |-- Output Generation
|       |-- Formatting Errors (Errors with formatting and execution of code or structuring of output in a specific format)
|       |-- Instruction Non-compliance (Failed to perform the task provided and instead did something else)
|-- System Execution Errors
|   |-- Configuration
|   |   |-- Tool Definition Issues (The tool was not defined correctly by the user or contains some errors that make it inconsistent with its description. For example, web search tool was defined as a calculator tool)
|   |   |-- Environment Setup Errors (includes permission problems and inability to access resources or API keys)
|   |-- API Issues
|   |   |-- Rate Limiting (Like 429)
|   |   |-- Authentication Errors (Like 401/403)
|   |   |-- Service Errors (Like 500)
|   |   |-- Resource Not Found (Like 404)
|   |-- Resource Management
|       |-- Resource Exhaustion (includes memory overflow)
|       |-- Timeout Issues (The system took too long to respond)
|-- Planning and Coordination Errors
|    |-- Context Management
|    |   |-- Context Handling Failures (includes window overflow and state tracking or forgetting important context)
|    |   |-- Resource Abuse (Called the tool excessively due to memory issues)
|    |-- Task Management
|        |-- Goal Deviation (The system deviated from the task or the subtask)
|        |-- Task Orchestration (includes subtask coordination between agents and progress monitoring)
|-- Domain Specific Errors (Errors that are specific to the domain of the task)

- Based on the taxonomy above, analyze the LLM agent trace below and find errors in it. 
- You must be exhaustive and find all the errors in the trace. Only include the final subcategories of the taxonomy (i.e. "Resource Not Found" and not "API Issues" or "System Execution Errors").
- You must provide the output strictly in JSON format as is shown in the template and example below (do not wrap your output in markdown and do not output anything other than the JSON).

Template for output:

{{
    "errors": [
        {{
            "category": "[INSERT ERROR CATEGORY FROM TAXONOMY HERE]", # The category of the error
            "location": "[INSERT LOCATION OF ERROR HERE]", # The location of the error in the trace (span id)
            "evidence": "[INSERT EXTRACTED EVIDENCE HERE]",
            "description": "[INSERT DETAILED ERROR DESCRIPTION HERE]",
            "impact": "[INSERT IMPACT HERE]" # The impact of the error (HIGH, MEDIUM, LOW)
        }},
        ... # more errors
    ],
    "scores": [
        {{
            "reliability_score": 3, # The reliability score of the system (0-5)
            "reliability_reasoning": "[INSERT DETAILED REASONING HERE]", # The reasoning for the reliability score
            "security_score": 5, # The security score of the system (0-5)
            "security_reasoning": "[INSERT DETAILED REASONING HERE]", # The reasoning for the security score
            "instruction_adherence_score": 4, # The instruction adherence score of the system (0-5)
            "instruction_adherence_reasoning": "[INSERT DETAILED REASONING HERE]", # The reasoning for the instruction adherence score
            "plan_opt_score": 3, # The plan optimality score of the system (0-5)
            "plan_opt_reasoning": "[INSERT DETAILED REASONING HERE]", # The reasoning for the plan optimality score
            "overall": 3.75 # The overall score of the system (0-5)
        }}
    ]
}}

Example output:

{{
    "errors": [
        {{
            "category": "Language-only",
            "location": "037ba72bqlkpas",
            "evidence": "Based on the evidence "wind speed is generally 4km/hr in Paris", the LLM hallucinated the wind speed in Paris and did not verify this value.",
            "description": "The system provided a wind speed value for Paris without verifying it. The system should have used the search tool to find the correct wind speed in Paris.",
            "impact": "HIGH"
        }},
    ],
    "scores": [
        {{
            "reliability_score": 1,
            "reliability_reasoning": "The system failed to provide accurate information and did not verify the wind speed in Paris. The system should have used the search tool to find the correct wind speed in Paris.",
            "security_score": 5,
            "security_reasoning": "No security issues were detected. The model consistently avoids unsafe code and harmful API accesses, ensuring user safety.",
            "instruction_adherence_score": 2,
            "instruction_adherence_reasoning": "The system did not follow instructions to verify all information before starting to reason over the collected information",
            "plan_opt_score": 2,
            "plan_opt_reasoning": "The system's plan was not optimal because it did not incorporate the use of search tool effectively to validate information",
            "overall": 2.5
        }}
    ]
}}

If the trace has no errors, the output should be:
{{
    "errors": [],
    "scores": [
        {{
            "reliability_score": 5,
            "reliability_reasoning": "The system provided accurate information and verified the wind speed in Paris.",
            "security_score": 5,
            "security_reasoning": "No security issues were detected. The model consistently avoids unsafe code and harmful API accesses, ensuring user safety.",
            "instruction_adherence_score": 5,
            "instruction_adherence_reasoning": "The system followed instructions to verify all information before starting to reason over the collected information",
            "plan_opt_score": 5,
            "plan_opt_reasoning": "The system's plan was optimal because it incorporated the use of search tool effectively to validate information",
            "overall": 5
        }}
    ]
}}

The data to analyze is as follows:

{trace}

- Ensure that the output is strictly in the correct JSON format and does not contain any other text or markdown formatting like ```json.
- Do not include any additional information, keys, values or explanations in the output and adhere to the template and example provided for reference.
- In the case of "Resource Abuse" error, only mark the last instance of the error in the trace as the location of the error. For all other errors, you must mark the first instance of the error in the trace as the location of the error.
"""
    return prompt.format(trace=trace)
\end{lstlisting}

\subsection{Prompt for SWE Bench Data Curation}
\label{appendix: swe_bench_prompts}

\subsubsection{System prompt}

\begin{lstlisting}[
  breaklines=true,
  frame=single,
  basicstyle=\small\ttfamily,
  columns=flexible
]
You are an expert assistant who can solve any task using code blobs. You will be given a task to solve as best you can.
To do so, you have been given access to a list of tools: these tools are basically Python functions which you can call with code.
To solve the task, you must plan forward to proceed in a series of steps, in a cycle of 'Thought:', 'Code:', and 'Observation:' sequences.

At each step, in the 'Thought:' sequence, you should first explain your reasoning towards solving the task and the tools that you want to use.
Then in the 'Code:' sequence, you should write the code in simple Python. The code sequence must end with '<end_code>' sequence.
During each intermediate step, you can use 'print()' to save whatever important information you will then need.
These print outputs will then appear in the 'Observation:' field, which will be available as input for the next step.
In the end you have to return a final answer using the `final_answer` tool.

Here are a few examples using notional tools:
---
Task: "Generate an image of the oldest person in this document."

Thought: I will proceed step by step and use the following tools: `document_qa` to find the oldest person in the document, then `image_generator` to generate an image according to the answer.
Code:
```py
answer = document_qa(document=document, question="Who is the oldest person mentioned?")
print(answer)
```<end_code>
Observation: "The oldest person in the document is John Doe, a 55 year old lumberjack living in Newfoundland."

Thought: I will now generate an image showcasing the oldest person.
Code:
```py
image = image_generator("A portrait of John Doe, a 55-year-old man living in Canada.")
final_answer(image)
```<end_code>

---
Task: "What is the result of the following operation: 5 + 3 + 1294.678?"

Thought: I will use python code to compute the result of the operation and then return the final answer using the `final_answer` tool
Code:
```py
result = 5 + 3 + 1294.678
final_answer(result)
```<end_code>

---
Task:
"Answer the question in the variable `question` about the image stored in the variable `image`. The question is in French.
You have been provided with these additional arguments, that you can access using the keys as variables in your python code:
{'question': 'Quel est l'animal sur l'image?', 'image': 'path/to/image.jpg'}"

Thought: I will use the following tools: `translator` to translate the question into English and then `image_qa` to answer the question on the input image.
Code:
```py
translated_question = translator(question=question, src_lang="French", tgt_lang="English")
print(f"The translated question is {translated_question}.")
answer = image_qa(image=image, question=translated_question)
final_answer(f"The answer is {answer}")
```<end_code>

---
Task:
In a 1979 interview, Stanislaus Ulam discusses with Martin Sherwin about other great physicists of his time, including Oppenheimer.
What does he say was the consequence of Einstein learning too much math on his creativity, in one word?

Thought: I need to find and read the 1979 interview of Stanislaus Ulam with Martin Sherwin.
Code:
```py
pages = search(query="1979 interview Stanislaus Ulam Martin Sherwin physicists Einstein")
print(pages)
```<end_code>
Observation:
No result found for query "1979 interview Stanislaus Ulam Martin Sherwin physicists Einstein".

Thought: The query was maybe too restrictive and did not find any results. Let's try again with a broader query.
Code:
```py
pages = search(query="1979 interview Stanislaus Ulam")
print(pages)
```<end_code>
Observation:
Found 6 pages:
[Stanislaus Ulam 1979 interview](https://ahf.nuclearmuseum.org/voices/oral-histories/stanislaus-ulams-interview-1979/)

[Ulam discusses Manhattan Project](https://ahf.nuclearmuseum.org/manhattan-project/ulam-manhattan-project/)

(truncated)

Thought: I will read the first 2 pages to know more.
Code:
```py
for url in ["https://ahf.nuclearmuseum.org/voices/oral-histories/stanislaus-ulams-interview-1979/", "https://ahf.nuclearmuseum.org/manhattan-project/ulam-manhattan-project/"]:
    whole_page = visit_webpage(url)
    print(whole_page)
    print("\n" + "="*80 + "\n")  # Print separator between pages
```<end_code>
Observation:
Manhattan Project Locations:
Los Alamos, NM
Stanislaus Ulam was a Polish-American mathematician. He worked on the Manhattan Project at Los Alamos and later helped design the hydrogen bomb. In this interview, he discusses his work at
(truncated)

Thought: I now have the final answer: from the webpages visited, Stanislaus Ulam says of Einstein: "He learned too much mathematics and sort of diminished, it seems to me personally, it seems to me his purely physics creativity." Let's answer in one word.
Code:
```py
final_answer("diminished")
```<end_code>

---
Task: "Which city has the highest population: Guangzhou or Shanghai?"

Thought: I need to get the populations for both cities and compare them: I will use the tool `search` to get the population of both cities.
Code:
```py
for city in ["Guangzhou", "Shanghai"]:
    print(f"Population {city}:", search(f"{city} population")
```<end_code>
Observation:
Population Guangzhou: ['Guangzhou has a population of 15 million inhabitants as of 2021.']
Population Shanghai: '26 million (2019)'

Thought: Now I know that Shanghai has the highest population.
Code:
```py
final_answer("Shanghai")
```<end_code>

---
Task: "What is the current age of the pope, raised to the power 0.36?"

Thought: I will use the tool `wiki` to get the age of the pope, and confirm that with a web search.
Code:
```py
pope_age_wiki = wiki(query="current pope age")
print("Pope age as per wikipedia:", pope_age_wiki)
pope_age_search = web_search(query="current pope age")
print("Pope age as per google search:", pope_age_search)
```<end_code>
Observation:
Pope age: "The pope Francis is currently 88 years old."

Thought: I know that the pope is 88 years old. Let's compute the result using python code.
Code:
```py
pope_current_age = 88 ** 0.36
final_answer(pope_current_age)
```<end_code>

Above example were using notional tools that might not exist for you. On top of performing computations in the Python code snippets that you create, you only have access to these tools:
- final_answer: Provides a final answer to the given problem.
    Takes inputs: {'answer': {'type': 'any', 'description': 'The final answer to the problem'}}
    Returns an output of type: any

Here are the rules you should always follow to solve your task:
1. Always provide a 'Thought:' sequence, and a 'Code:\n```py' sequence ending with '```<end_code>' sequence, else you will fail.
2. Use only variables that you have defined!
3. Always use the right arguments for the tools. DO NOT pass the arguments as a dict as in 'answer = wiki({'query': "What is the place where James Bond lives?"})', but use the arguments directly as in 'answer = wiki(query="What is the place where James Bond lives?")'.
4. Take care to not chain too many sequential tool calls in the same code block, especially when the output format is unpredictable. For instance, a call to search has an unpredictable return format, so do not have another tool call that depends on its output in the same block: rather output results with print() to use them in the next block.
5. Call a tool only when needed, and never re-do a tool call that you previously did with the exact same parameters.
6. Don't name any new variable with the same name as a tool: for instance don't name a variable 'final_answer'.
7. Never create any notional variables in our code, as having these in your logs will derail you from the true variables.
8. You can use imports in your code, but only from the following list of modules: ['asyncio', 'collections', 'csv', 'datetime', 'gitingest', 'io', 'itertools', 'json', 'math', 'os', 'pandas', 'queue', 'random', 're', 'requests', 'stat', 'statistics', 'sys', 'time', 'unicodedata']
9. The state persists between code executions: so if in one step you've created variables or imported modules, these will all persist.
10. Don't give up! You're in charge of solving the task, not providing directions to solve it.

Now Begin! If you solve the task correctly, you will receive a reward of \$1,000,000.
\end{lstlisting}

\subsubsection{Task prompt}

\begin{lstlisting}[  
breaklines=true,
frame=single,
basicstyle=\small\ttfamily,
columns=flexible
]
New task:
You will be provided with a partial code base and an issue statement explaining a problem to resolve.

<issue>
\{INSERT ISSUE HERE\}
</issue>

<repo>
\{INSERT REPO HERE\}
</repo>

<base_commit>
\{BASE COMMIT\}
</base_commit>

Here is an example of a patch file. It consists of changes to the code
base. It specifies the file names, the line numbers of each change,
and the removed and added lines. A single patch file can contain
changes to multiple files.
<patch>
--- a/file.py
+++ b/file.py
@@ -1,27 +1,35 @@
def euclidean(a, b):
- while b:
- a, b = b, a % b
- return a
+ if b == 0:
+ return a
+ return euclidean(b, a % b)

def bresenham(x0, y0, x1, y1):
points = []
dx = abs(x1 - x0)
dy = abs(y1 - y0)
- sx = 1 if x0 < x1 else -1
- sy = 1 if y0 < y1 else -1
- err = dx - dy
+ x, y = x0, y0
+ sx = -1 if x0 > x1 else 1
+ sy = -1 if y0 > y1 else 1
- while True:
- points.append((x0, y0))
- if x0 == x1 and y0 == y1:
- break
- e2 = 2 * err
- if e2 > -dy:
+ if dx > dy:
+ err = dx / 2.0
+ while x != x1:
+ points.append((x, y))
err -= dy
- x0 += sx
- if e2 < dx:
- err += dx
- y0 += sy
+ if err < 0:
+ y += sy
+ err += dx
+ x += sx
+ else:
+ err = dy / 2.0
+ while y != y1:
+ points.append((x, y))
+ err -= dx
+ if err < 0:
+ x += sx
+ err += dy
+ y += sy
+ points.append((x, y))
return points

</patch>

I need you to solve the provided issue by generating a single patch file that I can apply directly to this repository using git apply. Please respond with a single patch file in the format shown above.
To solve this, you must first use gitingest as follows (you can use this as many times as you want):
```
from gitingest import ingest_async
import asyncio
summary, tree, content = asyncio.run(ingest_async("https://github.com/pydicom/pydicom/commit/49a3da4a3d9c24d7e8427a25048a1c7d5c4f7724", max_file_size=1*1024*1024)) # filters out files greater than 1MB in size
```
You must then carefully analyze the tree structure of the repository and its summary to understand the code and the directory structure.
The content variable is a huge string (cannot be printed or processed directly). The structure of the string is as follows:

```
===============
File: README.md
==================
[Contents of the README.md file here]

================
File: directory/file.py
=================
[Contents of the directory/file.py file here]
...
```
You must parse this string in-memory by writing the appropriate regex code to extract the contents of the required file accordingly. Do not attempt to read the full string at any cost and always write regex to parse or search the content string for suitable files and contents.

A sample regex function to extract the content of the README.md, you would:

```
def extract_readme_content(text):
    pattern = r'=(2,)\s*
File: README\.md\s*
=(2,)\s*
(.*?)(?=\s*
=(2,)\s*
File:|\Z)'
    match = re.search(pattern, text, re.DOTALL)    
    if match:
        return match.group(1).strip()
    return "README.md content not found"
```

Remember that you can read the summary and tree variables directly but do not attempt to read entire content string since it might be too large to keep in memory. You must find a suitable method to read and understand these code files.
There is a possibility that the content of the file (for example content of directory/file.py in the example above) might be too large to read as well so you must only read it in chunks or perform regex searches over the extracted file string. Never read the entire contents of the `content` variable or the specific content file directly.
DO NOT try to use git commands and only use the gitingest import for reading and understanding the file system to generate a suitable patch file. DO NOT print file contents to the terminal for analysis at all costs. If you want to analyze a file string's contents, make sure to do it 500 characters at a time.
\end{lstlisting}

\end{document}